\documentclass{article}

\usepackage[dblblindworkshop, final]{neurips_2026}
\workshoptitle{AI for Stochastic Dynamics}

\usepackage[utf8]{inputenc} % allow utf-8 input
\usepackage[T1]{fontenc}    % use 8-bit T1 fonts
\usepackage{hyperref}       % hyperlinks
\hypersetup{hidelinks}       % Safely removes boxes without option clashes
\usepackage{url}            % simple URL typesetting
\usepackage{booktabs}       % professional-quality tables
\usepackage{amsfonts}       % blackboard math symbols
\usepackage{nicefrac}       % compact symbols for 1/2, etc.
\usepackage{microtype}      % microtypography
\usepackage{xcolor}         % colors
\usepackage[pdftex]{graphicx}
\usepackage{amsmath}
\usepackage{bm}

\title{Direct Intermediate Initialization for Tilted Diffusion Samplers}

\author{%
  Gregory D. Bellchambers \\
  PhysicsX Ltd.\\
  London, EC2A 4DQ \\
  \texttt{greg.bellchambers@physicsx.ai} \\
}

\begin{document}

\maketitle

\begin{abstract}
Some diffusion posterior samplers construct Gaussian-tilted intermediate
distributions along the reverse process. We observe that these targets can be
pulled back to clean-space posteriors with weaker conditioning, with samples
transported analytically to the corresponding noisy-space target through a
Gaussian bridge. For the sequential Monte Carlo (SMC) sampler MCGDiff, the effective observation variance of this
pulled-back problem is up to twice the diffusion-noise variance. We exploit this structure to initialize MCGDiff directly
at an intermediate time: an approximate solver samples the softened clean-space
posterior, the Gaussian bridge maps these samples to the tilted target, and only
the remaining SMC suffix is run. This trades asymptotic consistency for
finite-particle performance. With moment-matching posterior sampling (MMPS) as the solver, the hybrid improves sliced
Wasserstein distance by roughly $2\times$ at matched particle count on a structured Gaussian-mixture inverse problem, and by more than an order of magnitude when
the posterior-relevant mode is rare under the prior. A prior-initialization control, which retains the bridge but drops the clean-space conditioning, shows that on MCGDiff's standard Gaussian-mixture benchmark most of the improvement is insensitive to the conditioning. Conditioning the initialization gives a further consistent gain on the structured problem, and becomes decisive on a rare-mode problem, where
resampling cannot repopulate a mode absent from the initial population.
\end{abstract}

\section{Introduction}

Diffusion generative models \citep{song2019generative, ho2020denoising, karras2022elucidating} provide expressive priors for
Bayesian inverse problems \citep{daras2024survey}. Given a pretrained unconditional model,
training-free posterior samplers typically incorporate observations either
through approximations to the posterior score \citep{dps, mmps} or through a sequence of
auxiliary distributions that gradually introduce the likelihood \citep{cardoso2024monte, dcps, pnpdm}. The latter
include particle and SMC \citep{doucetfg01} methods such as MCGDiff \citep{cardoso2024monte}. Although SMC methods are
asymptotically consistent under appropriate assumptions, finite particle
populations can suffer from weight degeneracy, ancestral collapse, and loss
of posterior modes.

We consider linear inverse problems
\begin{equation}
\label{eq:y}
y=Ax_0 + \sigma_y \epsilon, \quad \epsilon \sim \mathcal{N}(0,I),
\end{equation}
with posterior $p(x_0|y)$. Several posterior samplers~\citep{cardoso2024monte, dcps, pnpdm, zhao2025generative} move through Gaussian tilts of diffusion marginals $p_t$,
\begin{equation}
\label{eq:vt}
    v_t(x_t)\propto p_t(x_t) \mathcal{N}(b_t; C_t x_t, R_t),
\end{equation}
which we call tilted intermediate targets. The pseudo-observation $b_t$, linear operator $C_t$, and covariance $R_t$ are designed so that the sequence terminates at the posterior of Eq~\ref{eq:y}, while the tilt constrains $x_0$ more weakly with increasing $t$. Each $v_t$ induces a clean-space posterior with weaker conditioning (Section~\ref{sec:direct-sampling}),
and our key observation is that this pullback can be substantially easier to sample than the
final posterior $p(x_0|y)$. We therefore combine an approximate posterior sampler with SMC by initializing the particle population directly at an intermediate time $t^\star$: we approximately
sample the pullback, transport those samples to $v_{t^\star}$ through an exact Gaussian
bridge, and run only the remaining SMC steps from $t^\star$ to $0$ (the \emph{suffix}), skipping those from the terminal time to $t^\star$ (the \emph{prefix}).

We apply this strategy to MCGDiff \citep{cardoso2024monte} with MMPS
\citep{mmps} as the clean-space sampler. At matched particle count,
the hybrid halves the sliced Wasserstein distance (SWD) of ESS-gated MCGDiff on a structured Gaussian mixture model (GMM) benchmark and improves on it about $15\times$ on a rare-mode problem.

Our first benchmark is the GMM problem introduced by MCGDiff and since used to compare against it \citep{dcps, DDSMC, boys2024tweedie}. Here, initializing from the prior instead of MMPS also beats MCGDiff given $16\times$ the particles, and still matches it at $8\times$ when MCGDiff starts from exact terminal samples (Appendix~\ref{app:sched}): the benchmark rewards prior coverage rather than fidelity to the conditioning. We therefore add a structured GMM and a rare-mode problem, the latter isolating conditioning fidelity from prior coverage.

\section{Method}
\label{sec:method}
Let $p_0(x_0)$ denote the data distribution represented by a pretrained diffusion model. In variance-preserving (VP) diffusion \citep{ho2020denoising, song2021scorebased} with $t \in [0, 1]$, the forward kernel is
\begin{equation}
\label{eq:vp-forward}
q_t(x_t|x_0) = \mathcal{N}\left(x_t; \sqrt{\bar{\alpha}_t}x_0, (1-\bar{\alpha}_t)I\right),
\end{equation}
where $\bar{\alpha}_0 = 1$ and $\bar{\alpha}_1 \approx 0$. The corresponding marginal is $p_t(x_t) = \int q_t(x_t|x_0) p_0(x_0)d x_0$.

We consider initializing an existing diffusion posterior sampler at an intermediate time $t^\star < 1$ by directly sampling the intermediate target $v_{t^\star}$ as an auxiliary inverse problem. When the intermediate is a Gaussian-tilted diffusion marginal, this can be done by sampling a softened inverse problem at clean-data time, followed by an analytic tilt-conditioned Gaussian bridge. We discuss computational cost in \ref{app:cost}.

\subsection{Direct sampling of a Gaussian-tilted diffusion marginal}
\label{sec:direct-sampling}

Recall the tilted intermediate target $v_t$ of Eq.~\ref{eq:vt}; it is the $x_t$-marginal of the joint
\begin{equation}
    \Pi_t(x_0,x_t)\propto p_0(x_0)q_t(x_t|x_0) \mathcal{N}(b_t; C_t x_t, R_t).
\end{equation}
Its $x_0$-marginal is a clean-space posterior with weaker conditioning than $v_t$:
\begin{equation}
\label{eq:x0-marginal}
    \tilde{\pi}_t(x_0)\propto p_0(x_0) \mathcal{N}(b_t; \sqrt{\bar{\alpha}_t} C_t x_0, R_t + (1-\bar{\alpha}_t)C_tC_t^T).
\end{equation}

We refer to $\tilde\pi_t$ as the pullback of the tilt. The conditional $\Pi_t(x_t | x_0)\propto q_t(x_t | x_0)\,\mathcal N(b_t;\,C_t x_t,\,R_t)$ is Gaussian and available in closed form. Drawing $x_0\sim\tilde\pi_t$ and then $x_t\sim\Pi_t(\cdot | x_0)$ gives $(x_0,x_t)\sim\Pi_t$, so $x_t\sim v_t$ exactly.

\subsection{Application to MCGDiff}

MCGDiff \citep{cardoso2024monte} constructs its intermediate Gaussian tilts in the singular-vector basis of the forward operator, with each measured direction becoming active once the diffusion noise exceeds a measurement-dependent threshold $\tau_i$. We restrict our initialization time to $t^\star>\max_i\tau_i$, so that all measured directions are active and the target is a single tilt of the form of Eq.~\ref{eq:vt}. Here $t^\star$ is a fractional position on MCGDiff's adapted grid (Appendix~\ref{app:sched}); $t$ increases with noise, so larger $t^\star$ gives a softer clean-space problem and a longer suffix. The $\tau_i$ lie on the same grid, and $\max_i \tau_i < 0.25$ in all benchmark runs; above this bound, $t^\star$ is a free hyperparameter. The corresponding $b_t$, $C_t$, and $R_t$, together with the exact pullback and bridge, are given in Appendix~\ref{app:mcgdiff-specialization}.

An important consequence of the pullback is that the clean-space inverse problem can be substantially softer than the original one. MCGDiff itself introduces Gaussian noise at approximately the diffusion noise level into its intermediate tilt; in the near-noiseless measurement limit, with unit singular values, the resulting clean-space likelihood therefore has effective variance $\sigma_{\mathrm{eff},t}^2 \simeq 2\sigma_t^2$, where $\sigma_t^2=\frac{1-\bar{\alpha}_t}{\bar{\alpha}_t}$. This makes the auxiliary inverse problem for MCGDiff particularly weakly conditioned.

We sample it with MMPS \citep{mmps}, which approximates the intractable likelihood score by
replacing $p(x_0|x_t)$ with a Gaussian matching its first two moments; the approximation
improves as the effective observation noise grows, and MMPS needs no guidance scale or other
problem-specific tuning. We transport the resulting samples through the exact Gaussian bridge to initialize $v_{t^\star}$, and then run the unchanged MCGDiff suffix. The choice of $t^\star$ therefore trades an increasingly easy MMPS problem against a longer remaining SMC trajectory. The hybrid does not inherit MCGDiff's asymptotic consistency, since the suffix is initialized with equal-weight approximate samples, but this could be restored by replacing MMPS with another asymptotically consistent method.

\section{Related work}
Our approach is related to recent methods that improve diffusion posterior sampling through reward- or
observation-informed initialization ($\Psi$-Sampler \citep{psi}, \citealt{chang2026provablediffusionposteriorsampling})
or easier auxiliary posterior distributions (DCPS \citep{dcps}, TGD \citep{tgd},
\citealt{zhao2025generative}). Unlike these, we directly initialize an intermediate target already defined by an
existing sampler, using its induced softened clean-space posterior and an exact bridge back
to the original diffusion-time target. See Appendix~\ref{app:related-work} for further discussion.

\begin{figure}[!t]
  \setlength{\textfloatsep}{8pt}
  \centering
  \includegraphics[width=\linewidth]{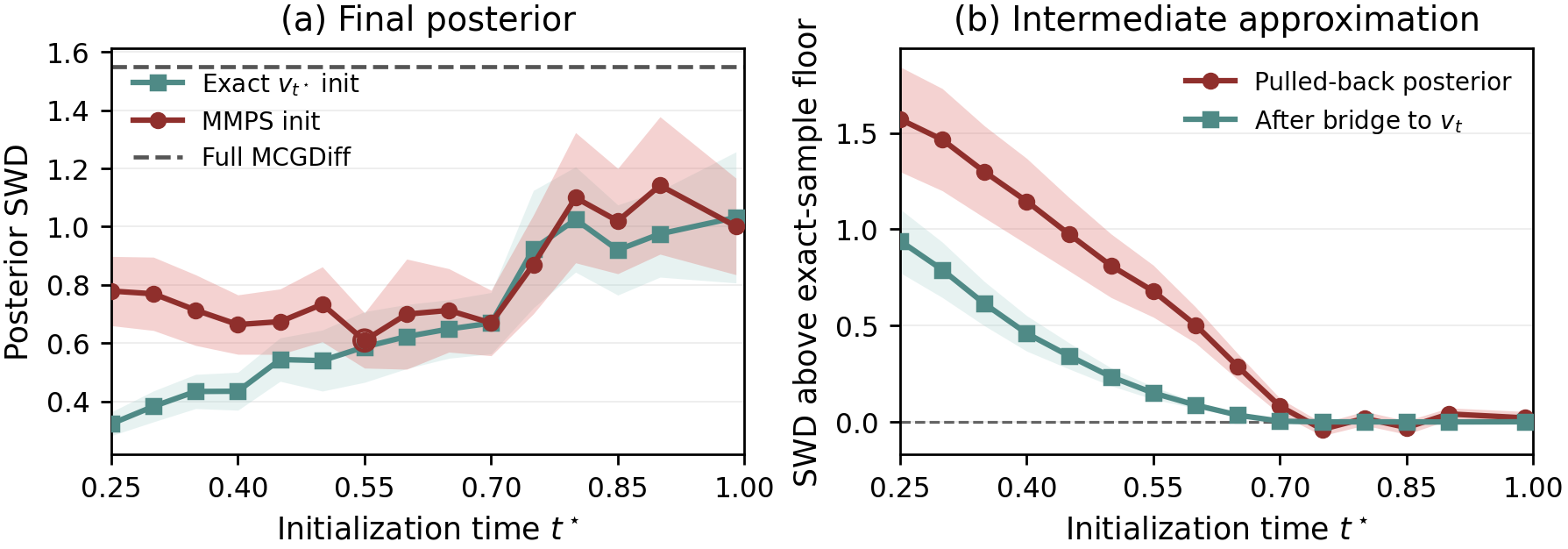}
  \caption{
Effect of initialization time $t^\star$ on the standard GMM benchmark.
(a) Final posterior SWD for oracle and MMPS initialization.
(b) MMPS approximation error before and after the exact bridge to $v_{t^\star}$. Generated with 1000 MMPS denoising steps and averaged over 80 seeds.
  }
  \label{fig:tstar}
\end{figure}

\section{Experiments}
We evaluate three inverse GMM problems for which both the exact posterior and MCGDiff intermediate targets \(v_t\) can be sampled analytically. This allows us to isolate the quality of intermediate initialization from the behavior of the subsequent SMC suffix. Performance is measured by sliced Wasserstein distance (SWD) to exact posterior samples; see Appendix~\ref{app:gmm-benchmarks} for full benchmark definitions. As also observed by \citet{zhao2025generative}, we find that MCGDiff improves when resampling is performed only when the particle ESS falls below \(0.7N\), rather than at every step. We therefore use this stronger ESS-gated variant for all full MCGDiff and initialized suffixes, while also reporting the original MCGDiff baseline. We use $N=1{,}000$ particles for MCGDiff and intermediate initialization runs, plus baselines with larger $N$ (up to $16{,}000$) for full MCGDiff and, on the standard GMM, for oracle initialization at $t^\star=1$.

The benchmarks separate two finite-particle failures. Loss of diversity can be repaired by a population with broad prior coverage; loss of support cannot, since resampling prunes particles but does not create them. We therefore include prior initialization at $t^\star$ as a control: it replaces the pullback posterior $\tilde{\pi}_{t^\star}$ with the prior $p_0$ in the clean-space step, and transports through the same exact bridge to $v_{t^\star}$. Both initializations bypass the prefix and enter the suffix with the observed directions conditioned on the measurement via the tilt; they differ only in whether the clean-space draw is conditioned. The standard and structured GMMs primarily probe lost diversity; the rare-mode benchmark isolates lost support, with $\rho$ setting the prior probability of the posterior-relevant mode and hence how likely a finite prior-drawn population is to miss it entirely.

Figure~\ref{fig:tstar}(a) compares oracle $v_{t^\star}$ initialization with MMPS initialization across values of $t^\star$ on the standard GMM benchmark. As $t^\star$ increases, the induced clean-space posterior becomes progressively softer and is sampled more accurately by MMPS (Figure~\ref{fig:tstar}(b)). This comes at the cost of a longer remaining MCGDiff suffix, so the final posterior SWD exhibits a broad trade-off between initialization accuracy and suffix length rather than a sharply tuned optimum. In contrast, oracle initialization improves as $t^\star$ decreases, as expected when increasingly large portions of the SMC trajectory are bypassed using exact intermediate samples. At the terminal time ($t^\star=1$), where no trajectory is bypassed, there is still substantial improvement with exact initialization of $v_{t^\star}$ over standard MCGDiff. This is not a consequence of bypassing the prefix, but of $v_{t^\star}$ correcting the Gaussian approximation to the terminal marginal, which is inaccurate under this benchmark's diffusion schedule (Appendix~\ref{app:sched}).

Across benchmarks, Table \ref{tab:gmm_appendix_mmps100} compares MMPS initialization with MCGDiff, direct MMPS, oracle initialization, and prior-based controls using 100 MMPS denoising steps. Direct MMPS on the original posterior is consistently weaker than most baselines, showing that the gain is not a result of replacing MCGDiff with MMPS, but of the hybrid.

On the standard GMM, both MMPS- and prior-initialization substantially improve on ESS-gated MCGDiff, with roughly $2\times$ and $3\times$ SWD improvement, respectively. Even with $16\times$ the population size allocated to the MCGDiff baseline, there is a clear benefit of intermediate initialization. Much of this margin is the terminal correction discussed above: MCGDiff initialized with exact terminal samples ($t^\star=1$) needs about $2\times$ the particles to match MMPS initialization and about $8\times$ to match prior initialization. Since prior initialization discards the conditioning of unobserved directions entirely over $[t^\star,1]$, and only weakly conditions observed directions via the bridge, this shows that MCGDiff's prefix returns a worse finite-particle approximation than a single biased step. Beyond the terminal-marginal correction, the gain continues to grow as $t^\star$ decreases: over the bypassed interval the tilt is weakly informative, so resampling incurs ancestral collapse in exchange for almost no selection signal. ESS-gated resampling improves on MCGDiff for the same reason. This identifies finite-particle degradation in the high-noise prefix of MCGDiff, which is precisely the portion our construction bypasses.

The structured GMM, inspired by \citet{zhao2025generative}, introduces random means, component-dependent covariances, and a shifted measurement. Here prior initialization remains useful, but MMPS initialization is better at every $t^\star$ below the terminal time (SWD half the MCGDiff baseline). Only at $16\times$ population size does MCGDiff outperform intermediate initialization. Finally, the rare-mode benchmark is deliberately constructed so that the observation strongly favors a component with very small prior probability. A finite prior-drawn population may therefore contain few or no particles from the posterior-relevant mode. In this setting, prior-based initialization and MCGDiff can fail because the relevant mode is never adequately represented, whereas MMPS initialization can populate that mode, despite direct MMPS performing poorly on this task. This benchmark gives the clearest advantage for MMPS initialization, which improves on MCGDiff by an order of magnitude.

Figures \ref{fig:posterior-samples-standard} and \ref{fig:posterior-samples-rare} show representative samples for all methods on the standard GMM and rare-mode GMM problems, respectively.

\section{Conclusion}
Gaussian-tilted diffusion marginals can be sampled indirectly: pull the tilt back to a softer clean-space posterior, sample it, and transport forward through an exact Gaussian bridge. Applied to MCGDiff, this bypasses the high-noise prefix, improving SWD by $2\times$ at matched particle count on a structured GMM and by more than an order of magnitude when the posterior-relevant mode is rare under the prior. Prior rather than posterior clean-space samples show that MCGDiff's standard GMM benchmark is largely insensitive to conditioning fidelity. The evaluation is confined to GMMs with one-dimensional observations; imaging and scientific inverse problems are the natural next steps. Our construction requires only that the intermediate tilt is Gaussian and linear in $x_t$; the original measurement model does not enter the pullback. Extending to non-linear problems with intermediate targets built from a measurement linearization \citep{enkg, blade} is a promising direction.

\bibliographystyle{plainnat}
\bibliography{biblio}

%%%%%%%%%%%%%%%%%%%%%%%%%%%%%%%%%%%%%%%%%%%%%%%%%%%%%%%%%%%%
\clearpage
\appendix
\part*{Appendix}

\section{Additional method details}
\subsection{Clean-space pullback and Gaussian bridge}
\label{app:gaussian-bridge}

The construction follows directly from standard linear-Gaussian
marginalization and conditioning identities
\citep{pml2Book}. Interpret the Gaussian tilt as an observation in reparameterized form,
\begin{equation}
    b_t = C_t x_t + \eta,
    \qquad
    \eta\sim\mathcal{N}(0,R_t).
\end{equation}
Substitute in the VP forward process, Eq. \ref{eq:vp-forward},
\begin{equation}
    x_t
    =
    \sqrt{\bar{\alpha}_t}x_0
    +
    \sqrt{1-\bar{\alpha}_t}\,\epsilon,
    \qquad
    \epsilon\sim\mathcal{N}(0,I),
\end{equation}
to give
\begin{equation}
    b_t\mid x_0
    \sim
    \mathcal{N}\!\left(
        \sqrt{\bar{\alpha}_t}C_t x_0,\,
        R_t+(1-\bar{\alpha}_t)C_tC_t^\top
    \right).
\end{equation}
Multiplying this likelihood by $p_0(x_0)$ gives $\tilde{\pi}_t$ in Eq~\ref{eq:x0-marginal}.

\subsection{Application to MCGDiff}
\label{app:mcgdiff-specialization}
Let $A=USV^\top$ and work in the singular-vector coordinates used by
MCGDiff \citep{cardoso2024monte}. For measured direction $i$, define
\begin{equation}
    \bar y_i = \frac{(U^\top y)_i}{s_i}.
\end{equation}
The direction becomes active at the MCGDiff threshold $\tau_i$, satisfying
\begin{equation}
    \bar{\alpha}_{\tau_i}\sigma_y^2
    =
    (1-\bar{\alpha}_{\tau_i})s_i^2 .
\end{equation}
For $t\geq\tau_i$, MCGDiff uses the Gaussian factor
\begin{equation}
    \mathcal{N}\!\left(
        \sqrt{\bar{\alpha}_t}\bar y_i;
        x_{t,i},
        \gamma_{t,i}
    \right),
    \qquad
    \gamma_{t,i}
    =
    1-(1-\kappa)
    \frac{\bar{\alpha}_t}{\bar{\alpha}_{\tau_i}} ,
\end{equation}
where $\kappa$ is MCGDiff's noise floor parameter.

Therefore, in the notation of Section~\ref{sec:direct-sampling},
$C_t$ selects the active singular-vector coordinates,
\begin{equation}
    b_{t,i}=\sqrt{\bar{\alpha}_t}\bar y_i,
    \qquad
    R_{t,ii}=\gamma_{t,i}.
\end{equation}

Applying Eq.~\eqref{eq:x0-marginal}, each active coordinate contributes the
clean-space likelihood
\begin{equation}
    \mathcal{N}\!\left(
        \bar y_i;
        x_{0,i},
        \sigma_{\mathrm{eff},t,i}^2
    \right),
    \qquad
    \sigma_{\mathrm{eff},t,i}^2
    =
    \frac{\gamma_{t,i}+1-\bar{\alpha}_t}
         {\bar{\alpha}_t}.
    \label{eq:mcgdiff-effective-noise}
\end{equation}
This is the softened inverse problem sampled by MMPS.

To transport a clean sample back to diffusion time $t$, note that for each
active coordinate
\begin{align}
    q_t(x_{t,i}\mid x_{0,i})
    &=
    \mathcal{N}\!\left(
        x_{t,i};
        \sqrt{\bar{\alpha}_t}x_{0,i},
        1-\bar{\alpha}_t
    \right),\\
    g_t(x_{t,i})
    &=
    \mathcal{N}\!\left(
        \sqrt{\bar{\alpha}_t}\bar y_i;
        x_{t,i},
        \gamma_{t,i}
    \right).
\end{align}
Their product is Gaussian in $x_{t,i}$, with
\begin{align}
    \sigma_{\mathrm{br},t,i}^2
    &=
    \left(
        \frac{1}{1-\bar{\alpha}_t}
        +
        \frac{1}{\gamma_{t,i}}
    \right)^{-1},\\
    \mu_{\mathrm{br},t,i}
    &=
    \sigma_{\mathrm{br},t,i}^2
    \left(
        \frac{\sqrt{\bar{\alpha}_t}x_{0,i}}
             {1-\bar{\alpha}_t}
        +
        \frac{\sqrt{\bar{\alpha}_t}\bar y_i}
             {\gamma_{t,i}}
    \right).
\end{align}
Inactive directions (unmeasured directions, together with measured directions with $t < \tau_i$) are sampled from the ordinary forward kernel; in our experiments $t^\star > \max_i\tau_i$, so only the former arise. Sampling these coordinate-wise Gaussian conditionals gives the exact bridge
used to initialize the MCGDiff suffix.

\subsection{Diffusion schedules}
\label{app:sched}
For MCGDiff we follow the schedule used by its GMM implementation. The
underlying VP discretization contains 1,000 time points, with the linear
$\beta$ schedule stored in decreasing order from $0.02$ to $10^{-4}$. We
note that, while the paper reports a maximum $\beta$ value of
$\beta_{\mathrm{max}} = 0.2$, the accompanying code uses
$\beta_{\mathrm{max}} = 0.02$, as do \citet{DDSMC} when replicating this
benchmark. We adopt $\beta_{\mathrm{max}} = 0.02$ to match the reference
implementation. The same applies to MCGDiff's
noise-floor parameter: we use the repository default $\kappa = 10^{-4}$,
whereas \citet{DDSMC} use $\kappa = 10^{-2}$. MCGDiff does not uniformly
subsample the raw diffusion indices, but uses a custom
measurement-dependent timestep-selection procedure
\citep{cardoso2024monte}. The resulting grid depends on the singular
values of the forward operator and the observation-noise variance. Our
reported $t^\star$ denotes a position in this adapted MCGDiff grid,
rather than a raw diffusion index. We use 100 diffusion steps for
MCGDiff throughout.

\paragraph{Consequence for terminal initialization.}
At $\beta_{\mathrm{max}} = 0.02$ we have $\bar\alpha_1 = 4.1 \times
10^{-5}$, so $\sqrt{\bar\alpha_1} \approx 6.4 \times 10^{-3}$. In the
standard GMM benchmark the outermost component means have $\lVert \mu
\rVert = 452$, giving a residual displacement
$\sqrt{\bar\alpha_1}\,\lVert \mu \rVert \approx 2.9$ against a noise
standard deviation of $1$ per coordinate. A Gaussian terminal
distribution is therefore a poor approximation, and an
$\mathcal{N}(0,I)$ initial population under-represents the outer
components before any SMC step is taken. The issue is specific to this benchmark:
$\mathcal{N}(0,I)$ is a good approximation to the terminal distribution
of the structured and rare-mode priors, whose means are smaller and
spread over more directions. This is reflected in
Table~\ref{tab:gmm_appendix_mmps100}, where oracle initialization at
$t^\star = 1.0$ improves on full MCGDiff by $1.7\times$ on the standard
GMM but only $1.1\times$ on the structured GMM. For the standard GMM we therefore also report oracle initialization at $t^\star = 1$ with larger $N$ as a terminal-corrected baseline.

We use $\eta=0.85$ for the DDIM \citep{ddim} steps of the MCGDiff suffix throughout. Repeating the standard GMM benchmark at
$\eta=1$ over the first 30 seeds and comparing to $\eta=0.85$ results over the same seeds shows that the comparisons are insensitive to $\eta$ (MCGDiff 1k: $2.383\to2.380$; with ESS gating: $1.958\to1.804$). For MMPS sampling of the clean-space posterior, we use $\eta=1$ throughout.

MMPS uses a separate auxiliary VP schedule with $\beta$ increasing
linearly from $10^{-4}$ to $0.2$ over 1,000 time points. For the
100-step MMPS experiments, we truncate this auxiliary schedule at the
first time for which $\bar{\alpha}_t \leq 10^{-8}$ and place 100
approximately uniformly spaced raw diffusion indices between zero and
this terminal time. This avoids spending steps in the extreme
high-noise tail, where the diffusion marginal is already effectively
Gaussian. For the 1,000-step MMPS experiments we instead use the full
auxiliary schedule.

The MMPS schedule is used only to numerically sample the pulled-back
clean-space posterior. The target $v_{t^\star}$ and the subsequent
Gaussian bridge are defined using the MCGDiff schedule.

\subsection{Computational cost}
\label{app:cost}
Table~\ref{tab:gmm_appendix_mmps1000} repeats the benchmark comparison with
1,000 rather than 100 MMPS steps and shows little change in performance,
suggesting that 100 MMPS steps are sufficient for these experiments. In a
practical application, where the score is represented by a neural network,
using 100 steps for both MMPS and MCGDiff with $t^\star = 0.5$ requires 100 diffusion steps for the initialization followed by 50 for the
MCGDiff suffix, against 100 for standard MCGDiff. The number of diffusion
steps therefore increases only moderately.

Per-step cost depends on the choice of clean-space sampler. MMPS requires
gradients through the score network, and is among the more expensive options;
in our GMM benchmarks its cost was negligible, since the score is analytic and
the observation is one-dimensional. It was chosen because it needs no
guidance-scale or other problem-specific tuning, which keeps the ablation of
the initialization mechanism clean. The pullback is a standard linear-Gaussian inverse problem, so
any approximate posterior sampler can be substituted, including cheaper samplers such as DPS \citep{dps}.

A natural alternative is simply to give MCGDiff more particles.
Table~\ref{tab:gmm_appendix_mmps100} reports $2\times$ and $16\times$
populations. On the standard GMM, even $16\times$ does not make MCGDiff with the $\mathcal N(0,I)$ terminal competitive with MMPS initialization, although an exact-terminal MCGDiff matches it at about $2\times$. On the structured benchmark, $16\times$
with ESS-gated resampling does close the gap. Particles are evaluated in
parallel, so in these experiments a larger population costs memory rather than
sequential depth. In high-dimensional problems this no longer holds: the
population will not fit in memory at once, and the extra sequential batches of
score-network evaluations erode much of the apparent advantage over the
initialization prefix.

\subsection{Sliced Wasserstein evaluation}
\label{app:swd}
We compute SW$_1$ between 1,000 samples from each approximate method and 1,000 samples from the exact posterior, using 1,000 shared random projections per inverse-problem instance. Methods run with larger particle populations are subsampled to 1,000 final particles only after inference so that all SWD comparisons use the same empirical sample size.

\section{GMM benchmark definitions}
\label{app:gmm-benchmarks}

We use three Gaussian mixture model (GMM) inverse problems. The first follows the benchmark construction used by MCGDiff, while the second is inspired by, but is not identical to, the structured GMM experiments of \cite{zhao2025generative}. The third is a deliberately constructed rare-mode problem designed to stress methods initialized from finite prior samples.

\paragraph{Standard GMM.}
Following MCGDiff, we use a $d=800$ dimensional, $25$-component unit-covariance GMM. For $i,j\in\{-2,-1,0,1,2\}$, the component means are
\begin{equation}
    \mu_{ij}
    =
    (8i,8j,\ldots,8i,8j)
    \in\mathbb{R}^{800},
\end{equation}
and each component has covariance $I$. The mixture weights are randomly drawn from a $\chi^2$ distribution once and shared by all 80 instances, which differ in $A$, $\sigma_y$ and $y$.

We only use one-dimensional linear observations. We first sample a Gaussian random matrix and retain its singular vectors, while replacing its singular values by sorted draws from $\mathcal{U}(0,1)$. An observation is generated from a prior sample $x_0$ according to
\begin{equation}
    y = A x_0 + \sigma_y \epsilon,
    \qquad
    \epsilon\sim\mathcal{N}(0,I),
\end{equation}
where $\sigma_y$ is sampled uniformly between zero and the largest singular value of $A$. No additional observation offset is used.

\paragraph{Structured GMM.}
Our structured benchmark is inspired by, but is not identical to, the GMM
experiment of \cite{zhao2025generative}. We use $d=800$ and $K=10$. The unnormalised
mixture weights are independently sampled as
\begin{equation}
    \tilde w_k \sim \chi^2_1,
    \qquad
    w_k = \frac{\tilde w_k}{\sum_j \tilde w_j}.
\end{equation}
Independently for each component,
\begin{equation}
    \mu_k \sim \operatorname{Unif}([-8,8]^d),
    \qquad
    \Lambda_k = \lambda_k\lambda_k^\top + I,
    \qquad
    \lambda_k \sim \operatorname{Unif}([0,1]^d).
\end{equation}
The observation operator $A\in\mathbb{R}^{1\times d}$ is a random unit
row vector. Following the outlier construction of \cite{zhao2025generative}, we choose
\begin{equation}
    y = \mathbb{E}_{x\sim p_0}[Ax] + \omega
      = A\mathbb{E}_{p_0}[x] + \omega,
    \qquad \omega=5,
\end{equation}
and define the posterior using a Gaussian likelihood with observation
standard deviation $\sigma_y=0.5$ (i.e.\ $R=0.25$). Unlike \cite{zhao2025generative},
we therefore use a fixed observation covariance and a unit-norm
one-dimensional forward operator. Since the prior is a GMM and the
likelihood is linear Gaussian, the posterior remains analytically tractable.

\paragraph{Rare-mode GMM.}
We construct a two-component GMM designed to test whether a finite particle
population can recover a posterior-relevant mode that is very rare under the
prior. We use $d=64$, identity covariance for both components, and mixture
weights
\begin{equation}
    (w_{\mathrm{common}},w_{\mathrm{rare}})
    =
    (1-\rho,\rho).
\end{equation}
We draw a random unit vector $a$, which defines the observed direction, and
a second random vector $q$ perpendicular to $a$, normalized so that
$\|q\|=\sqrt{d}$. Since $q$ is orthogonal to $a$, displacement along $q$
is completely invisible to the measurement. A prior sample transported through the bridge is therefore corrected along $a$ while its position along $q$ is unchanged, so it can satisfy the measurement while remaining in the wrong component.

The component means are
\begin{equation}
    \mu_{\mathrm{common}}=-4a-4q,
    \qquad
    \mu_{\mathrm{rare}}=+4a+4q.
\end{equation}
We use
\begin{equation}
    A=a^\top,
    \qquad
    y=4,
    \qquad
    \sigma_y=0.5.
\end{equation}
Thus the measurement strongly favours the rare component,
since $A\mu_{\mathrm{common}}=-4$ while
$A\mu_{\mathrm{rare}}=4=y$, even though that component has very small
prior probability. The additional separation along $q$ makes the two
components well separated in the full space while remaining invisible
to the observation. We vary $\rho$ to control how poorly the rare component
is represented by a finite prior sample.

\section{Related work (extended)}
\label{app:related-work}
Several recent methods improve posterior sampling by replacing an unconditional high-noise initialization with one informed by the observation or reward. $\Psi$-Sampler \citep{psi} samples a reward-aware terminal distribution before running SMC, while \citealt{chang2026provablediffusionposteriorsampling} warm-start the conditional reverse diffusion with a Monte Carlo posterior-score estimator. These methods are close in spirit to our direct initialization, but
initialize at the terminal time rather than at an intermediate target.

A second line of work exploits easier intermediate posterior problems. DCPS \citep{dcps} recursively decomposes posterior sampling into shorter diffusion-time subproblems with annealed likelihoods, while TGD \citep{tgd} defines an explicit annealing path of clean-space tempered posteriors and uses diffusion states to move between them. \cite{zhao2025generative} instead proposes alternative Gaussian tilts to improve SMC under informative likelihoods. Our method differs in that we do not design a new bridging family: we directly initialize an existing SMC pathway at one of its own intermediate targets, thereby skipping the prefix. We derive the clean-space posterior implied by that target and the exact bridge back to it; because this pullback is more weakly conditioned, the tilted target can be populated with approximately independent MMPS samples before running only the remaining SMC suffix.

Other methods condition in clean space and transport forward. DAPS \citep{daps} reconstructs a clean sample from the current state, conditions it on the observation, and renoises through the forward kernel. DDSMC \citep{DDSMC} places this step inside an SMC algorithm. Both approximate the intractable noisy likelihood with a point estimate of the clean sample, so their intermediate targets approximate the noisy-space posterior rather than being auxiliary tilts, and there is no equivalent exact pullback posterior.

Tilted transport \citep{tiltedtransport} maps a linear-Gaussian posterior to a different, more strongly tilted ``boosted'' posterior that can be easier to sample, before transporting samples back to the target. Our construction moves in the opposite direction: an existing noisy-space tilt is pulled back to a more weakly conditioned clean-space posterior. MCGDiff \citep{cardoso2024monte} itself uses a related transport identity, converting noisy observations into noiseless constraints at intermediate diffusion times, but does not exploit the resulting clean-space softening considered here.

\section{Additional figures and tables}
\clearpage

\begin{figure}[t]
  \centering
  \includegraphics[width=\linewidth]{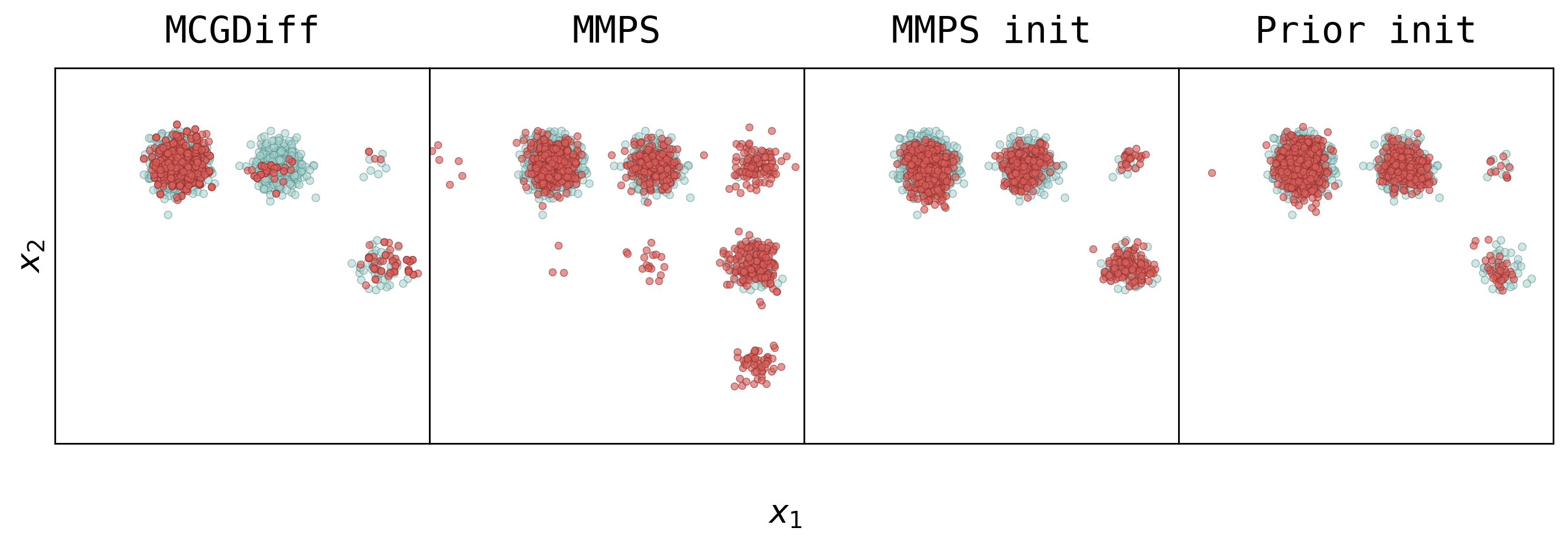}
  \caption{
    Posterior samples on the standard GMM benchmark for MCGDiff, direct MMPS, MMPS initialization, and prior initialization.
    MMPS and Prior initialization give qualitatively comparable samples on this relatively simple benchmark. Teal: exact posterior samples; Red: approximate method samples.
  }
  \label{fig:posterior-samples-standard}
\end{figure}

\begin{figure}[t]
  \centering
  \includegraphics[width=\linewidth]{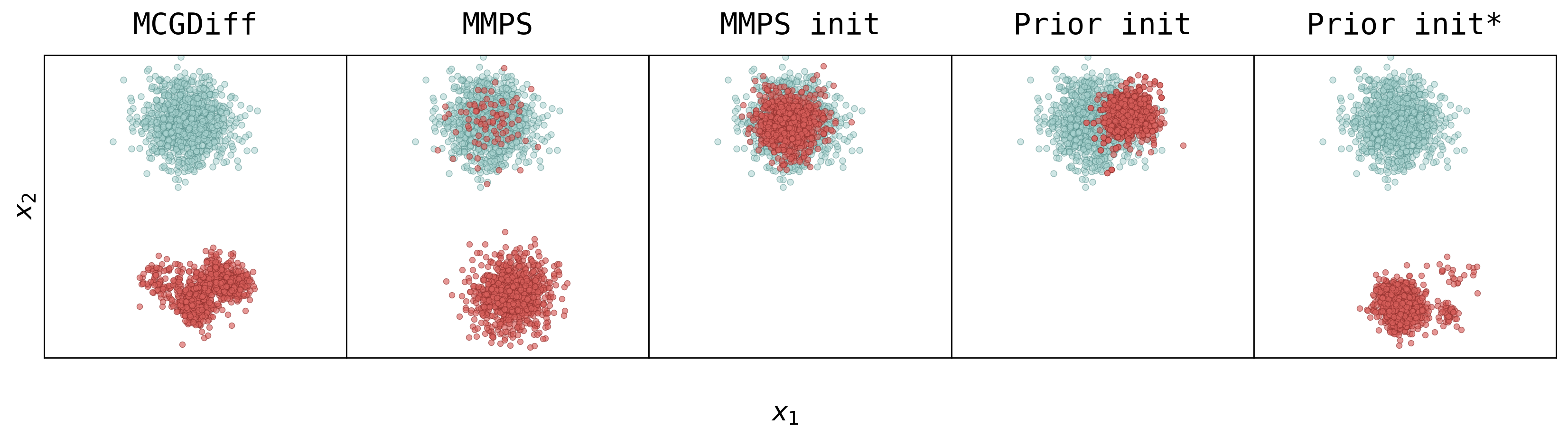}
  \caption{
    Posterior samples on the rare-mode GMM benchmark for MCGDiff, direct MMPS, MMPS initialization, and two separate prior initialization examples.
    Unless marked otherwise, initialized methods use $t^\star=0.3$ and ESS-thresholded resampling at $0.7N$.
    Prior initialization$^\star$ instead uses $t^\star=0.4$ with resampling at every step. This variant illustrates a failure mode of prior-based initialization: it collapses to the wrong posterior mode, whereas the ESS-thresholded prior initialization recovers the correct mode but remains less accurate than MMPS initialization. Teal: exact posterior samples; Red: approximate method samples.
  }
  \label{fig:posterior-samples-rare}
\end{figure}

\begin{table*}[t]
\caption{Sliced Wasserstein distance (lower is better) for the GMM benchmarks using MMPS initialization with 100 denoising steps. Values are mean $\pm$ 95\% confidence interval over 80 random seeds. Initialization methods use the ESS$<0.7$ MCGDiff suffix. Column $N$ gives the number of MCGDiff particles used. Among approximate initialization methods, the best result is bold and the second best is underlined. ``Exact posterior floor'' is the SWD calculated from comparing two batches of 1000 samples of exact posterior samples. Oracle init values for $t^\star=1$ are included for higher particles $N$ for the standard GMM benchmark to provide a fairer baseline, removing the bias due to the benchmark's inaccurate terminal initialization (see Appendix~\ref{app:sched}).}
\centering
\footnotesize
\setlength{\tabcolsep}{3.5pt}
\begin{tabular}{lllrrrr}
\toprule
Method & $N$ & $t^\star$ & GMM & Structured & Rare ($\rho=0.002$) & Rare ($\rho=0.01$) \\
\midrule
\multicolumn{7}{l}{\textit{Baselines}} \\
MCGDiff & 1k & -- & $2.307 \pm 0.45$ & $1.271 \pm 0.20$ & $5.818 \pm 0.40$ & $3.052 \pm 0.67$ \\
MCGDiff (ESS 0.7) & 1k & -- & $1.856 \pm 0.40$ & $0.762 \pm 0.12$ & $2.836 \pm 0.66$ & $0.280 \pm 0.14$ \\
MCGDiff & 2k  & -- & $1.897 \pm 0.31$ & $0.838 \pm 0.12$ & -- & -- \\
MCGDiff (ESS 0.7) & 2k & -- & $1.519 \pm 0.31$ & $0.548 \pm 0.09$ & -- & -- \\
MCGDiff & 16k & -- & $1.354 \pm 0.23$ & $0.382 \pm 0.05$ & -- & -- \\
MCGDiff (ESS 0.7) & 16k & -- & $1.326 \pm 0.26$ & $0.310 \pm 0.05$ & -- & -- \\
\addlinespace[2pt]
\multicolumn{7}{l}{\textit{Oracle init}} \\
 & 1k & 0.25 & $0.395 \pm 0.06$ & $0.279 \pm 0.03$ & $0.080 \pm 0.00$ & $0.078 \pm 0.00$ \\
 & 1k & 0.30 & $0.446 \pm 0.08$ & $0.337 \pm 0.05$ & $0.086 \pm 0.00$ & $0.079 \pm 0.00$ \\
 & 1k & 0.40 & $0.488 \pm 0.07$ & $0.404 \pm 0.07$ & $0.125 \pm 0.00$ & $0.093 \pm 0.00$ \\
 & 1k & 0.60 & $0.777 \pm 0.19$ & $0.532 \pm 0.09$ & -- & -- \\
 & 1k & 1.00 & $1.090 \pm 0.24$ & $0.693 \pm 0.11$ & -- & -- \\
 & 2k & 1.00 & $0.877 \pm 0.15$ & -- & -- & -- \\
 & 4k & 1.00 & $0.765 \pm 0.15$ & -- & -- & -- \\
 & 8k & 1.00 & $0.641 \pm 0.13$ & -- & -- & -- \\
 & 12k & 1.00 & $0.571 \pm 0.14$ & -- & -- & -- \\
 & 16k & 1.00 & $0.570 \pm 0.14$ & -- & -- & -- \\
\addlinespace[2pt]
\multicolumn{7}{l}{\textit{MMPS init}} \\
 & 1k & 0.25 & $1.059 \pm 0.17$ & $0.417 \pm 0.05$ & $\mathbf{0.180 \pm 0.00}$ & $\mathbf{0.111 \pm 0.00}$ \\
 & 1k & 0.30 & $0.920 \pm 0.16$ & $\mathbf{0.382 \pm 0.05}$ & $\underline{0.191 \pm 0.00}$ & $\underline{0.115 \pm 0.00}$ \\
 & 1k & 0.40 & $0.899 \pm 0.17$ & $\underline{0.412 \pm 0.06}$ & $0.225 \pm 0.01$ & $0.125 \pm 0.00$ \\
 & 1k & 0.60 & $0.963 \pm 0.18$ & $0.507 \pm 0.07$ & -- & -- \\
 & 1k & 1.00 & $0.959 \pm 0.17$ & $0.735 \pm 0.13$ & -- & -- \\
\addlinespace[2pt]
\multicolumn{7}{l}{\textit{Prior init (exact)}} \\
 & 1k & 0.25 & $\underline{0.670 \pm 0.10}$ & $0.461 \pm 0.06$ & $1.256 \pm 0.43$ & $0.260 \pm 0.01$ \\
 & 1k & 0.30 & $0.675 \pm 0.10$ & $0.461 \pm 0.07$ & $1.231 \pm 0.43$ & $0.244 \pm 0.01$ \\
 & 1k & 0.40 & $\mathbf{0.610 \pm 0.11}$ & $0.515 \pm 0.08$ & $1.180 \pm 0.43$ & $0.232 \pm 0.01$ \\
 & 1k & 0.60 & $0.849 \pm 0.16$ & $0.546 \pm 0.10$ & -- & -- \\
 & 1k & 1.00 & $1.155 \pm 0.20$ & $0.719 \pm 0.13$ & -- & -- \\
\addlinespace[2pt]
\multicolumn{7}{l}{\textit{Additional references}} \\
Direct MMPS & 1k & -- & $3.216 \pm 0.41$ & $1.668 \pm 0.19$ & $5.879 \pm 0.03$ & $4.261 \pm 0.03$ \\
Prior & 1k & -- & $7.339 \pm 0.51$ & $3.037 \pm 0.16$ & $6.445 \pm 0.04$ & $6.394 \pm 0.03$ \\
Exact posterior floor & 1k & -- & $0.240 \pm 0.03$ & $0.135 \pm 0.01$ & $0.057 \pm 0.00$ & $0.057 \pm 0.00$ \\
\bottomrule
\end{tabular}
\label{tab:gmm_appendix_mmps100}
\end{table*}

\begin{table*}[t]
\caption{Sliced Wasserstein distance (lower is better) for the GMM benchmarks using MMPS with 1000 denoising steps. Values are mean $\pm$ 95\% confidence interval over 80 random seeds. Methods use the ESS$<0.7$ MCGDiff suffix unless marked otherwise. The number of MCGDiff particles used for all relevant methods is 1,000. Among approximate methods, the best result is bold and the second best is underlined. Data in this table generated with different random seeds to corresponding data in Figure~\ref{fig:tstar}.}
\centering
\footnotesize
\setlength{\tabcolsep}{3.5pt}
\begin{tabular}{llrr}
\toprule
Method & $t^\star$ & GMM & Structured \\
\midrule
\multicolumn{4}{l}{\textit{Baselines}} \\
MCGDiff (1k) & -- & $2.615 \pm 0.46$ & $1.271 \pm 0.20$ \\
MCGDiff (1k, ESS 0.7) & -- & $1.784 \pm 0.29$ & $0.762 \pm 0.12$ \\
\addlinespace[2pt]
\multicolumn{4}{l}{\textit{Oracle init}} \\
 & 0.25 & $0.444 \pm 0.07$ & $0.320 \pm 0.06$ \\
 & 0.30 & $0.466 \pm 0.08$ & $0.322 \pm 0.04$ \\
 & 0.40 & $0.556 \pm 0.10$ & $0.377 \pm 0.06$ \\
 & 0.60 & $0.665 \pm 0.11$ & $0.485 \pm 0.07$ \\
 & 1.00 & $1.206 \pm 0.23$ & $0.679 \pm 0.10$ \\
\addlinespace[2pt]
\multicolumn{4}{l}{\textit{MMPS init}} \\
 & 0.25 & $1.063 \pm 0.19$ & $\mathbf{0.385 \pm 0.05}$ \\
 & 0.30 & $1.013 \pm 0.19$ & $0.426 \pm 0.06$ \\
 & 0.40 & $0.867 \pm 0.16$ & $\underline{0.391 \pm 0.06}$ \\
 & 0.60 & $0.782 \pm 0.17$ & $0.527 \pm 0.09$ \\
 & 1.00 & $1.164 \pm 0.22$ & $0.695 \pm 0.12$ \\
\addlinespace[2pt]
\multicolumn{4}{l}{\textit{Prior init (exact)}} \\
 & 0.25 & $0.820 \pm 0.15$ & $0.456 \pm 0.07$ \\
 & 0.30 & $\underline{0.644 \pm 0.10}$ & $0.485 \pm 0.08$ \\
 & 0.40 & $\mathbf{0.615 \pm 0.10}$ & $0.487 \pm 0.07$ \\
 & 0.60 & $0.833 \pm 0.21$ & $0.506 \pm 0.08$ \\
 & 1.00 & $1.174 \pm 0.22$ & $0.634 \pm 0.10$ \\
\addlinespace[2pt]
\multicolumn{4}{l}{\textit{Additional references}} \\
Direct MMPS & -- & $3.267 \pm 0.42$ & $1.668 \pm 0.19$ \\
Prior & -- & $7.293 \pm 0.52$ & $3.037 \pm 0.16$ \\
Exact posterior floor & -- & $0.240 \pm 0.03$ & $0.135 \pm 0.01$ \\
\bottomrule
\end{tabular}
\label{tab:gmm_appendix_mmps1000}
\end{table*}

\end{document}